\documentclass[10pt,twocolumn,letterpaper]{article}

\usepackage{cvpr}
\usepackage{times}
\usepackage{epsfig}
\usepackage{graphicx}
\usepackage{amsmath}
\usepackage{amssymb}

\usepackage{subfigure}
\usepackage{tabularx}
\usepackage{multirow}

\newcommand{\CKM}{\checkmark}

\usepackage[breaklinks=true,bookmarks=false]{hyperref}

\cvprfinalcopy 

\def\cvprPaperID{****} 

\begin{document}
\newcolumntype{C}[1]{>{\centering\arraybackslash}p{#1}}

\title{SSA-CNN: Semantic Self-Attention CNN for Pedestrian Detection}

\author{Chengju Zhou\\
Nanyang Technological University, Singapore \\
\and
Meiqing Wu\\
Nanyang Technological University, Singapore 
\and
Siew-Kei Lam\\
Nanyang Technological University, Singapore 
}

\maketitle

\begin{abstract}

 Pedestrian detection plays an important role in many applications such as autonomous driving. We propose a  method that explores semantic segmentation results as self-attention cues to significantly improve the pedestrian detection performance. Specifically, a multi-task network is designed to jointly learn semantic segmentation and pedestrian detection from image datasets with weak box-wise annotations. The semantic segmentation feature maps are concatenated with corresponding convolution features maps to provide more discriminative features for pedestrian detection and pedestrian classification. By jointly learning segmentation and detection, our proposed pedestrian self-attention mechanism can effectively identify pedestrian regions and suppress backgrounds. In addition, we propose to incorporate semantic attention information from multi-scale layers into deep convolution neural network to boost pedestrian detection. Experiment results show that the proposed method achieves the best detection performance with MR of 6.27\% on Caltech dataset and obtain competitive performance on CityPersons dataset while maintaining high computational efficiency.

\end{abstract}

\section{Introduction}


Pedestrian detection is an essential task in diverse computer vision applications such as autonomous driving, surveillance and robotics \cite{geiger2012we}. Although significant research efforts in pedestrian detection have been undertaken in recent years with the renaissance of deep learning \cite{brazil2017illuminating,zhang2018occluded}, the state-of-the-art pedestrian detection performance still cannot match human perception \cite{zhang2016far}. The detection performance often suffers in challenging cases such as occlusion, blur, shape variations, etc. To address these challenges, mechanisms for handling specific cases have been proposed. These include using segmentation as prior for detection \cite{fidler2013bottom,hariharan2014simultaneous} and designing models for different pedestrian occlusion patterns \cite{tian2015deep,wang2012discriminative,zhou2017multi}. However, the above methods do not lend themselves well in real-world scenarios. For instance, \cite{fidler2013bottom} requires additional processing for the segmentation task, which relies on fine grain annotations and has high computational complexity. The methods in \cite{tian2015deep,wang2012discriminative,zhou2017multi} cannot cater to all occlusion patterns and often induce high computational complexity when exploiting individual model for specific occlusion pattern. FasterRCNN+ATT \cite{zhang2018occluded} exploits channel-wise attention for occluded pedestrian detection. However, their method needs additional efforts to obtain attention information from other datasets and has a high overall computational complexity. SDS-RCNN \cite{brazil2017illuminating} presents a framework for joint supervision on pedestrian detection and semantic segmentation. However, the SDS-RCNN only adds semantic segmentation branch to infuse semantic features into the backbone network layers and does not use the semantic segmentation results directly for pedestrian detection.


In this paper, we propose a self-attention mechanism which forces the detector to focus on the pedestrian-likely regions and suppresses the background regions. This self-attention mechanism is motivated by the fact that the pedestrian regions are illuminated in the semantic segmentation results.
The semantic segmentation results provide pixel-wise class information  which improves the inter-class discrimination ability for pedestrian classification and reduces the difficulty of pedestrian bounding box regression.
 Our proposed method is based on widely used object detection framework, Faster R-CNN \cite{ren2015faster}. A  multi-scale multi-task learning framework is proposed for semantic segmentation and pedestrian detection in both RPN and R-CNN stages. Specifically, two semantic segmentation branches are connected to network layers with different scales in the RPN and R-CNN network in order to obtain multi-scale semantic feature maps. Then the multi-scale semantic feature maps are used as semantic self-attention cues and concatenated with corresponding convolution feature maps as features for pedestrian detection in RPN and pedestrian classification in R-CNN respectively. Experiments on well-known pedestrian detection datasets show that the proposed method achieves considerable improvement over state-of-the-art methods.

The rest of the paper is organized as follows. A review of existing methods in pedestrian detection is presented in Section \ref{related_work}. Section \ref{proposed_method} introduces the proposed Semantic Self-Attention CNN (SSA-CNN). Section \ref{experiments} presents the experiment results on the Caltech and CityPersons datasets to demonstrate the superiority of the proposed approach over state-of-the-art methods. We conclude the paper in Section \ref{conclusion}.

\begin{figure*}[th]
\centering	\includegraphics[width=0.95\linewidth]{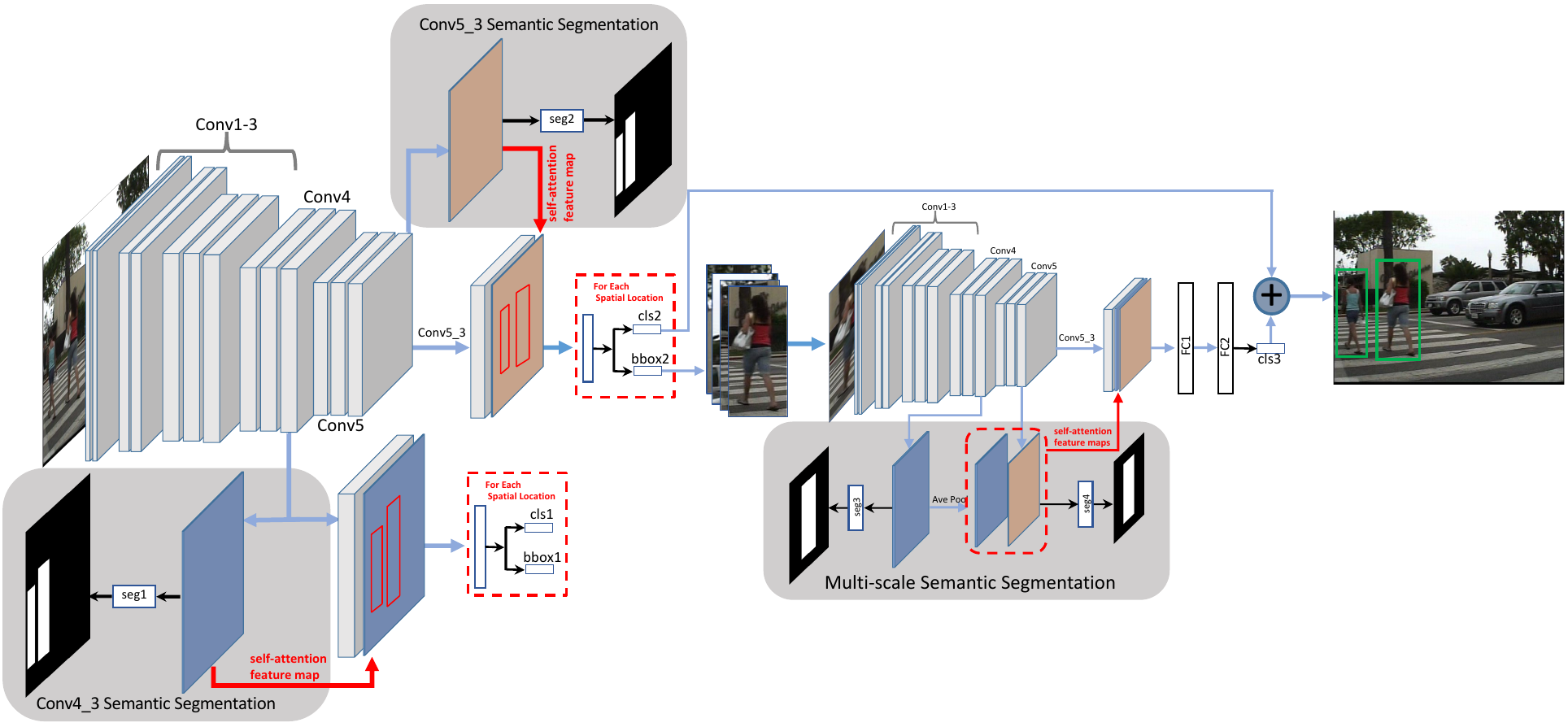}
\caption{Overview of proposed framework. The framework consists of two stages: Semantic Self-Attention RPN (referred as SSA-RPN) and Semantic Self-Attention R-CNN (referred as SSA-RCNN). In each stage, multi-scale semantic segmentation branches are added to conv4\_3 and conv5\_3 layers of VGG-16 network to generate semantic segmentation results as self-attention feature maps. The self-attention feature maps are then concatenated with corresponding convolution feature maps to work as features for pedestrian detection in SSA-RPN and pedestrian classification in SSA-RCNN. The detection results are obtained by applying Non-Maximum Suppress (NMS) on candidates with combined confidence scores from \textit{cls2} of SSA-RPN and \textit{cls3} of SSA-RCNN. The detection and segmentation branches connected to conv4\_3 layer in SSA-RPN are not used during inference.
}
\label{framework}
\end{figure*}

\section{Related Work}
\label{related_work}

Pedestrian detection has attracted a lot of attention and achieved notable progress in recent years \cite{benenson2014ten,zhang2016far}. Approaches for pedestrian detection have evolved rapidly from  Deformable Part Models (DPM) \cite{yan2014fastest,xu2014domain,felzenszwalb2010object,felzenszwalb2010cascade,girshick2012rigid} to Filtered Channel Feature (FCF) \cite{dollar2014fast,dollar2009integral,zhang2015filtered,zhang2016far,zhou2017fast,zhougroup}. Currently, deep convolution neural networks (DCNN) are the dominant approaches for pedestrian detection.

The Faster R-CNN \cite{ren2015faster} variants obtain top performance on several benchmarks including INRIA \cite{dalal2005histograms}, Caltech \cite{dollar2012pedestrian} and KITTI \cite{geiger2012we}. Faster R-CNN can be decomposed into two stages: Region Proposal Network (RPN) and Region-based Convolution Neural Network (R-CNN). The RPN aims to obtain object proposals through the objectness of candidates from each spatial location of an image with pre-defined aspect ratios and scales. The object proposals are fed into the classification sub-network R-CNN to determine the accurate localization and class it belongs to.
In Multi Scale CNN (MS-CNN) \cite{cai2016unified}, a unified DCNN is proposed to perform detection at various intermediate network layers such that the receptive fields match objects at different scales. By integrating the RPN of Faster R-CNN and a boosted forest as downstream classifier, the RPN+BF \cite{zhang2016faster} achieves miss rate (MR) 9.58\% when evaluating on Caltech reasonable setting \cite{dollar2012pedestrian}. The F-DNN (Fused Deep Neural Network) \cite{du2017fused} further reduces the MR to 8.65\% by combining more classification networks (including GoogleNet \cite{szegedy2015going} and ResNet-50 \cite{he2016deep}) as downstream classifier. With an individual semantic segmentation network trained on Cityscapes dataset \cite{cordts2016cityscapes}, the F-DNN+SS and F-DNN2+SS \cite{du2018fused} achieve MR 8.18\% and MR 7.67\% on Caltech reasonable setting. PCN \cite{wang2018pcn} utilizes two sub-networks which detects the pedestrians through body parts semantic information and context information respectively, and achieves MR 8.45\% on Caltech dataset. In order to detect small-scale pedestrians, TLL+TFA \cite{song2018small} proposes a method that integrates somatic  topological  line  localization  (TLL)  and  temporal feature  aggregation. TLL+TFA achieves MR 7.40\% on Caltech dataset and obtains better detection performance for pedestrians that are far away from camera.

Even though F-DNN2+SS achieves considerable improvement on detection performance, the main drawback is that it induces extreme high computational complexity due to the employment of several deep convolution neural networks. It is reported that F-DNN2+SS \cite{du2018fused} requires 2.48 seconds to process one image from Caltech dataset on NVIDIA Titan X GPU which is unacceptable for real-world scenarios such as autonomous driving and robotics. GDFL \cite{lin2018graininess} proposes to exploit scale-aware pedestrian attention masks and a zoom-in-zoom-out module to improve the capability of the feature maps to identify small and occluded pedestrians.  GDFL achieves MR 7.85\% on Caltech dataset with runtime of 0.05 seconds on single GPU.
Even though GDFL can run much faster than F-DNN2+SS, it has lower detection performance. In order to improve detection performance without compromising on computational efﬁciency, SDS-RCNN \cite{brazil2017illuminating} proposes to jointly learn the task of pedestrian detection and semantic segmentation by adding semantic segmentation branches to top network layers. SDS-RCNN achieves the best detection performance with MR 7.36\% on Caltech reasonable setting. Since SDS-RCNN infuses semantic features during training only, the network inference efficiency is unaffected and hence, it is faster than F-DNN2+SS.


The attention mechanism has been widely explored with DCNN in computer vision applications such as human pose estimation \cite{newell2016stacked}, image classification \cite{hu2017squeeze} and object detection \cite{bell2016inside,zhang2018occluded}. In SENet \cite{hu2017squeeze}, a novel architectural unit termed as “Squeeze-and-Excitation” (SE) block is proposed which adaptively recalibrates channel-wise feature responses by explicitly modelling interdependencies between channels. SENet can produce considerable performance improvement for image classification with minor additional computational cost. Inspired by \cite{hu2017squeeze}, FasterRCNN+ATT \cite{zhang2018occluded} proposes to employ channel-wise attention to handle occlusions for pedestrian detection. This is motivated by the findings that many channel features are localizable and often correspond to different body parts. An attention vector is learned from attention network to re-weight the top convolution channels as attention guidance and notable performance improvement is achieved for occluded cases.

\subsection{Contributions}
Our proposed method differs from the existing techniques in the following ways. 
Unlike FasterRCNN+ATT \cite{zhang2018occluded}, which exploit attentions to re-weight convolution channels, we propose to concatenate semantic feature maps as additional channels with corresponding network layers. These semantic feature maps can serve as semantic self-attention to boost the pedestrian detection in RPN and classification in R-CNN respectively. Our semantic feature maps are derived from a multi-task learning framework which simultaneously learns pedestrian detection and semantic segmentation. The proposed method also differs from SDS-RCNN \cite{brazil2017illuminating}, which adds a semantic segmentation branch to top network layer. Instead, we perform semantic segmentation from multiple network layers with different resolutions that integrates various granularity of semantic information into shared feature maps. The proposed method only requires box-wise annotations for the task of semantic segmentation while FasterRCNN+ATT \cite{zhang2018occluded} needs additional efforts or dataset (i.e., MPII Pose Dataset \cite{insafutdinov2016deepercut}) to design the occlusion patterns or train a human part detection network. Compared with SDS-RCNN, which only learns detection and segmentation from single-scale by sharing feature maps in backbone network through infused layer \cite{brazil2017illuminating}, the proposed method integrates the semantic features and detection features and further exploits multi-scale semantic feature maps as semantic self-attention cues to boost pedestrian detection.  


Our contributions can be summarized as follows:
\begin{enumerate}
\item[1)] We introduce a semantic self-attention mechanism to explore semantic segmentation results to boost pedestrian detection. The proposed attention mechanism only needs box-wise annotations to obtain the semantic 
information rather than pixel-wise annotations for the task of semantic segmentation.
\item[2)] We propose a multi-scale multi-task learning framework that jointly learns pedestrian detection and semantic segmentation from multi-scale intermediate network layers which can integrate different granularities semantic information into shared feature maps.
\item[3)] We achieve top detection performance on Caltech pedestrian dataset and competitive performance on CityPersons dataset while maintaining computational efficiency.
\end{enumerate}

\section{Proposed Method}
\label{proposed_method}

The proposed method extends the framework of Faster R-CNN, and it consists of two stages: Semantic Self-Attention RPN (referred as SSA-RPN) to generate pedestrian proposals and Semantic Self-Attention R-CNN (referred as SSA-RCNN) to refine the outputs from SSA-RPN. In SSA-RPN stage, the semantic feature maps derived from multi-scale  network layers  are concatenated with corresponding convolution feature maps of backbone network (i.e., VGG-16 \cite{simonyan2014very}) as features for pedestrian detection. In SSA-RCNN stage, the semantic feature maps obtained from multi-scale network layers are firstly pooled into same size and then concatenated with top convolution feature maps (i.e., conv5\_3) as features for pedestrian classification. In the rest of the paper, we will refer to the proposed method as SSA-CNN (Semantic Self-Attention CNN). Compared with prior work in \cite{brazil2017illuminating}, the proposed method explores multi-scale semantic segmentations to integrate multi-scale semantic information into shared feature maps and exploits semantic segmentation results as self-attention cues to boost pedestrian detection in RPN and pedestrian classification in R-CNN respectively. The proposed framework is illustrated in Fig. \ref{framework}.

\begin{figure*}[t]
\centering
\subfigure[]{ 
\includegraphics[width=1\linewidth]{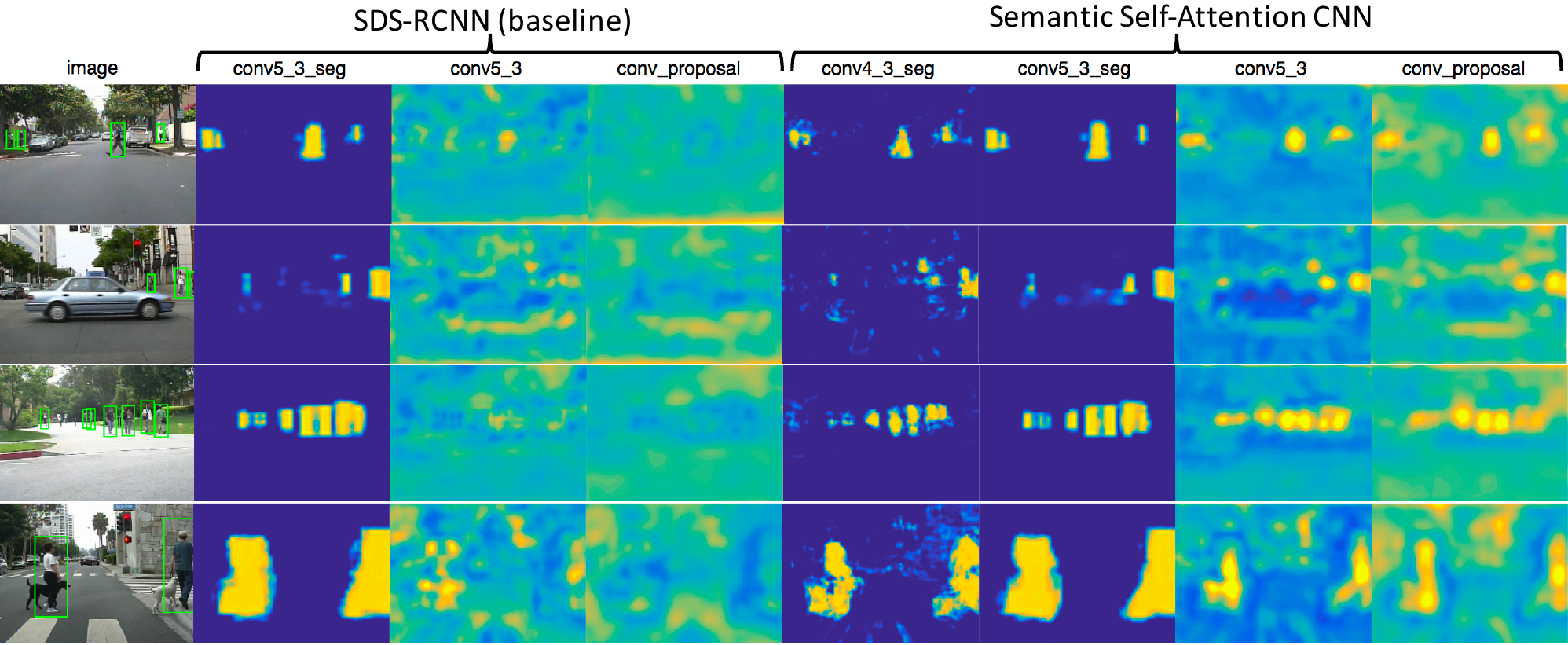}}  
\caption[]{Feature map visualization of conv5\_3\_seg, conv5\_3 and conv\_proposal layer in RPN stage of SDS-RCNN and proposed SSA-CNN. The conv4\_3\_seg feature maps from proposed method are also illustrated. Best viewed in color. We use the released model from SDS-RCNN to extract feature maps.}

\label{vis_cmp}
\end{figure*}


\subsection{Semantic Self-Attention RPN (SSA-RPN)}

The original RPN in Faster R-CNN aims to obtain a set of bounding box proposals with certain confidence levels for pedestrians by using a sliding window detector over each spatial location of input image with pre-defined scales and aspect ratio. 
The VGG-16 \cite{simonyan2014very} is used as the backbone network and we only keep the conv1-5 layers in the proposed semantic self-attention RPN. 

\begin{figure}[t]
\centering
\subfigure[]{ 
\includegraphics[width=1\linewidth]{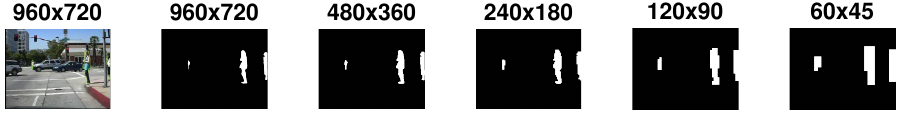}}
\subfigure[]{ 
\includegraphics[width=1\linewidth]{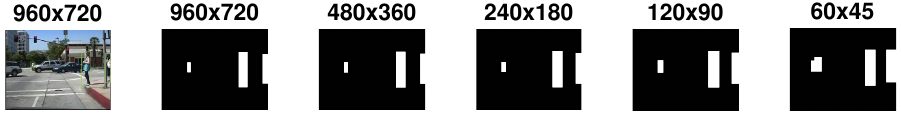}}   
\caption{Visualization of a) pixel-wise annotation  and b) box-wise annotation  at several scales used for SSA-RPN in proposed framework on Caltech dataset \cite{dollar2012pedestrian}. The titles indicate the resolution of corresponding scales. }
\label{rpn_anno_show}
\end{figure}

In order to obtain semantic feature maps, we add two semantic segmentation branches to conv4\_3 and conv5\_3 layers, which we call conv4\_3\_seg and conv5\_3\_seg respectively as highlighted in right part of Fig. \ref{framework}. The segmentation branches aim to integrate the semantic information into backbone network layers by exploiting the minor difference between box-wise annotations and pixel-wise annotations \cite{brazil2017illuminating} when image is downsampled significantly across the network layers as shown in Fig. \ref{rpn_anno_show}. 
 After obtaining the semantic feature maps, we concatenate them with corresponding convolution feature maps as features for pedestrian detection. In particular, the conv4\_3\_seg feature map is concatenated with conv4\_3 convolution feature map  while conv5\_3\_seg feature map is concatenated with conv5\_3 convolution feature map as shown in Fig. \ref{framework}.  The connection (highlighted in \textcolor{red}{red} color) between semantic segmentation results and combined feature maps for pedestrian detection in Fig. \ref{framework} fuses detection features and semantic segmentation features. This is different from SDS-RCNN \cite{brazil2017illuminating} that only has connection from top network layer to semantic segmentation results. As such, SDS-RCNN does not infuse detection features into the task of semantic segmentation.
 The influence of different connections of proposed method and SDS-RCNN can be observed from the visualization of RPN feature maps shown in Fig. \ref{vis_cmp}.  The conv5\_3 and its successive conv\_proposal feature maps of proposed SSA-RPN highlight the potential pedestrian regions while SDS-RCNN  fails to locate the pedestrian regions. The detection and segmentation branches attached to conv4\_3 layer are only used during training and therefore do not affect the network inference efficiency. The SSA-RPN can be trained by minimizing the following loss function:
\begin{equation}
L_{rpn} = \alpha_{c1}L_{c1} + \beta_{r1} L_{r1} + \gamma_{s1} L_{s1} + 
          \alpha_{c2}L_{c2} + \beta_{r2} L_{r2} + \gamma_{s2} L_{s2} 
\label{eq1}
\end{equation}
where $L_{c*}$, $L_{r*}$ and $L_{s*}$ are cross-entropy loss for classification, smooth-$L_1$ loss for bounding box regression and cross-entropy loss for semantic segmentation respectively. The $\alpha_{*}$, $\beta_{*}$ and $\gamma_{*}$ are weight of corresponding loss function. The classification loss  $L_{c*}$ and segmentation loss $L_{s*}$ are designed for binary classification problem (pedestrian vs. non-pedestrian). Following \cite{ren2015faster}, a proposal box is labelled as positive (i.e., pedestrian) if the intersection over union (IoU) with groundtruth box is larger than 0.5, and otherwise the box is considered as negative (i.e., non-pedestrian). The detection results of proposed semantic self-attention RPN are obtained after applying NMS with threshold 0.5 on the proposals.

Caffe \cite{jia2014caffe} is used to train the proposed SSA-RPN. We use SGD as optimizer with learning rate of 0.001, monument of 0.9 and we drop the learning rate by a factor of 10 every 70000 iterations. The mini-batch for training SSA-RPN is set to 1 ( one image in each iteration). Following the setting in \cite{brazil2017illuminating}, 120 proposals are randomly sampled with a ratio of 1:5 for pedestrian and non-pedestrian proposals from each image. The conv1-5 layers of SSA-RPN network are initialized from VGG-16 model trained on ImageNet \cite{deng2009imagenet} and other layer's parameters are initialized from a Gaussian distribution with standard deviation of 0.01. The parameters in conv1 layers are fixed in our training process. 

\begin{figure}[t]
\centering
\subfigure[]{ 
\includegraphics[width=1\linewidth]{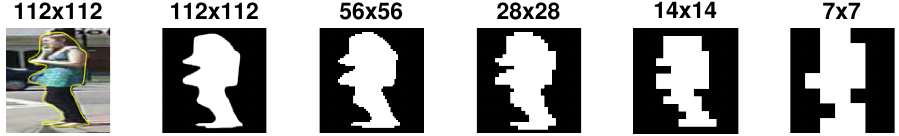}}
\subfigure[]{ 
\includegraphics[width=1\linewidth]{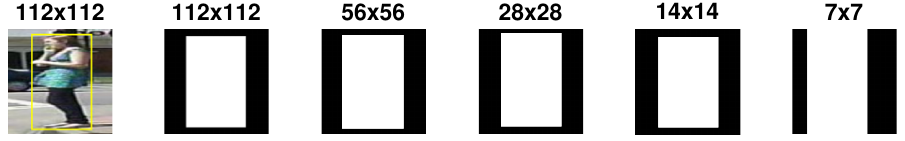}}   
\caption{Visualization of a) pixel-wise annotation  and b) box-wise annotation  at several scales used for SSA-RCNN in proposed framework on Caltech dataset \cite{dollar2012pedestrian}. The titles indicate the resolution of corresponding scales. Please note that the proposals from SSA-RPN are firstly added 25\% padding  and then resized into 112$\times$112 before being fed into proposed SSA-RCNN.}
\label{rcnn_anno_show}
\end{figure}

\subsection{Semantic Self-Attention R-CNN (SSA-RCNN)}

The goal of SSA-RCNN in the proposed framework is to further improve the detection performance by classifying the proposals from SSA-RPN into pedestrian or non-pedestrian. In Faster R-CNN for general object detection \cite{ren2015faster}, a ROI pooling layer is exploited to extract fixed dimension features for each proposal from the last convolution layer of backbone network in RPN (i.e., conv5\_3 in VGG-16). However, experiences in \cite{zhang2016faster} show that the pooling bins collapse if ROI's input resolution is smaller than output (i.e., 7 $\times$ 7 which is $112 \times 112$ in input image) which induces the extracted features flat and less discriminative. This problem becomes more severe in pedestrian detection since about 88\% pedestrians in Caltech dataset \cite{dollar2012pedestrian} and 64\% pedestrians in CityPersons dataset \cite{Shanshan2017CVPR} are lower than 112 pixels tall. To alleviate this issue, we crop the proposals from the RGB input image rather than from the top convolution network layer as shown in Fig. \ref{framework}. We add 25\% padding when cropping the proposals from input image and then resize the proposals to $112 \times 112$ before being fed into proposed SSA-RCNN as illustrated in Fig. \ref{rcnn_anno_show}. We use conv1-5 layers of VGG-16 as backbone network for SSA-RCNN.

In the proposed SSA-RCNN, we add two semantic segmentation branches to conv4\_3 and conv5\_3 layers respectively. In contrast to directly concatenating semantic feature maps with corresponding convolution feature maps in SSA-RPN, the semantic feature map obtained from conv4\_3 layer is pooled with stride 2 and then concatenated with semantic feature map from conv5\_3 layer as self-attention feature maps as shown in Fig. \ref{framework}. The conv5\_3 convolution feature map is then concatenated with the combined self-attention feature maps to serve as features for pedestrian classification. Compared with SDS-RCNN \cite{brazil2017illuminating} that only add semantic segmentation branch to conv5\_3 layer, we connect one more semantic segmentation branch to conv4\_3 layer and use combined semantic feature maps as semantic self-attention to boost pedestrian classification. Since we only use SSA-RCNN as binary classifier for pedestrian and non-pedestrian, the semantic self-attention R-CNN can be trained by minimizing the following loss function:
\begin{equation}
L_{rcnn} = \alpha_{c3}L_{c3}  + \gamma_{s3} L_{s3} + \gamma_{s4} L_{s4} 
\label{eq1}
\end{equation}
where $L_{c3}$, $L_{s3}$ and $L_{s4}$ are cross-entropy loss function for classification and semantic segmentation respectively. The $\alpha_{c3}$, $\gamma_{s3}$ and $\gamma_{s4}$ are weight of corresponding loss function.

We use Caffe to train the proposed semantic self-attention R-CNN. The SGD optimizer with learning rate 0.002 and monument 0.9 is employed. The learning rate is dropped by a factor of 10 for every 60000 iterations. The mini-batch is 25 when training proposed R-CNN and we define a proposal as pedestrian when its IoU with ground truth is larger than 0.8. This is motivated by \cite{brazil2017illuminating} which showed that a higher IoU threshold can provide better training samples. The parameters of conv1-5 layers are initialized from the trained SSA-RPN while others are initialized from Gaussian distribution with standard deviation 0.01. We fix the parameters in conv1 layers when training semantic self-attention R-CNN.

\begin{figure}[t]
\centering
\subfigure[]{ 
\includegraphics[width=0.45\linewidth]{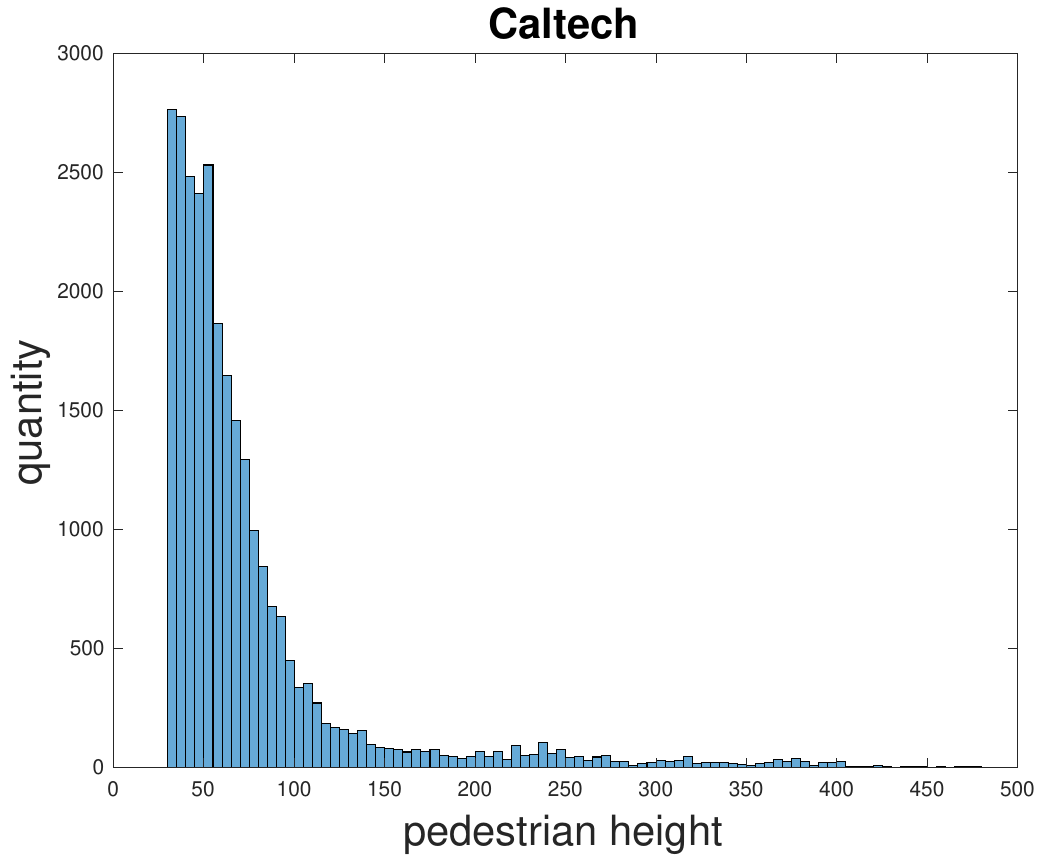}}
\subfigure[]{ 
\includegraphics[width=0.45\linewidth]{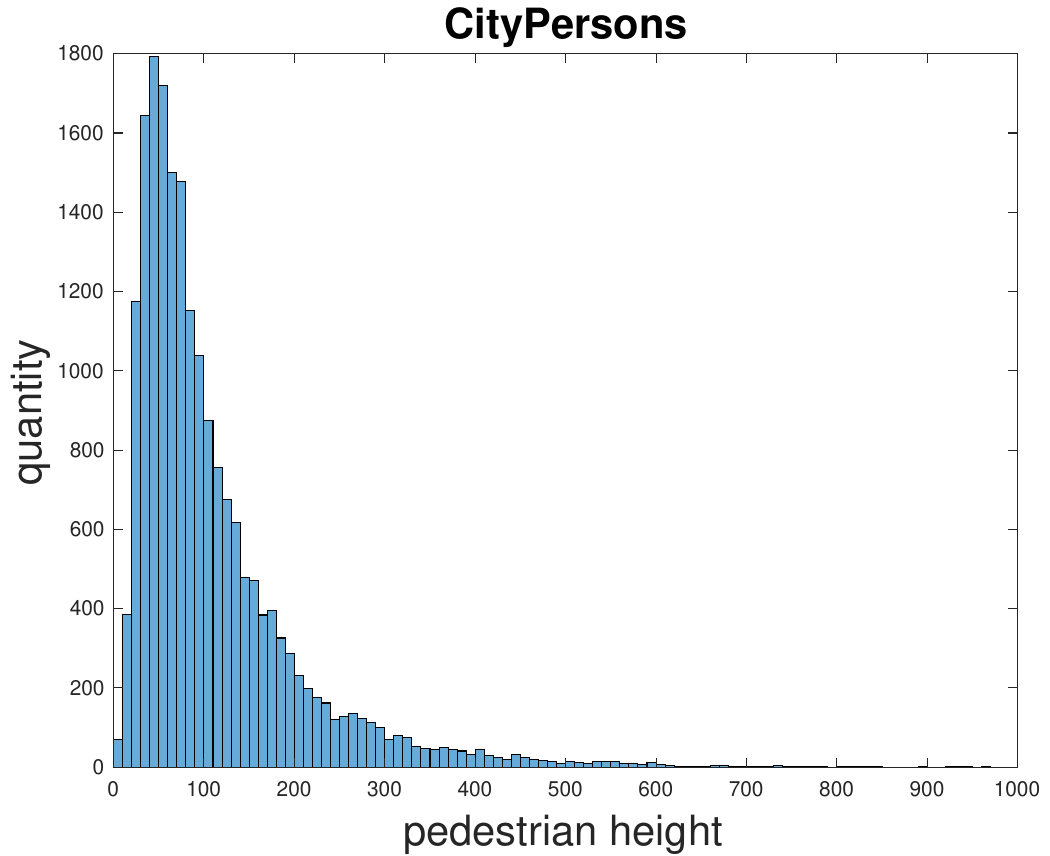}}   
\caption{Statistics of pedestrian height (pixels in original images) of Caltech and CityPersons  (training subset only).}
\label{ped_height_stat}
\end{figure}

\section{Experiments}
\label{experiments}

In this section, we first introduce the pedestrian detection datasets and evaluation metrics used in experiments. Then we show experiment results to compare the performance and runtime of the proposed SSA-CNN with state-of-the-art methods. Finally, we will report the results of our ablation studies for the proposed method on Caltech dataset.

\subsection{Datasets and Evaluation metrics}

We conduct experiments on two public datasets: Caltech \cite{dollar2012pedestrian} and CityPersons \cite{Shanshan2017CVPR}. Caltech and CityPersons are widely used benchmarks for the task of pedestrian detection.  For the Caltech dataset, we follow the approach in \cite{zhang2015filtered} wherein the training data is augmented by extracting one of every 3 frames instead of every 30 frames from the raw videos in \cite{dollar2014fast}. We refer to the former as Caltech10x. There are 42782 images with resolution of $640 \times 480$ in Caltech10x training set. The Caltech test set has 4024 images which includes 1014 positive images. The images are upsampled to $960 \times 720$ in our experiments. The CityPersons dataset is built upon the Cityscapes dataset \cite{cordts2016cityscapes} where data is collected from multiple cities and countries across Europe. There are a large number of occluded pedestrians in CityPersons dataset that makes it ideal one for evaluating the occlusion robustness of detection approaches.  The image resolution is $2048 \times 1024$ and we use original resolution in our experiments. We conduct experiments on original training and validation subset which includes 2975 and 500 images respectively. 
The statistics of pedestrian height of Caltech and CityPersons are shown in Fig. \ref{ped_height_stat}. It can be observed that CityPersons dataset has a large variations on pedestrian's height and includes more smaller pedestrians (i.e., pedestrians with less than 30 pixels tall) than Caltech dataset.

For the Caltech and CityPersons datasets, we use standard log-average miss rate (MR) between $10^{-2}$ and $10^{0}$ of false positive per image (FPPI) to evaluate the detection performance. The Reasonable setup is used for Caltech and CityPersons datasets where the pedestrian is at least 65\% visible and at least 50 pixels tall. In addition to the widely-used Reasonable setup, the Partial occlusion setup in which the occlusion ratio of pedestrian is from 1\% to 35\% is also used to evaluate detection performance on Caltech dataset. For CityPersons dataset, Small (visibility is between 20\% and 65\% with between 50 and 75 pixels tall), Heavy occlusion ( visibility is between 20\% and 65\% with at least 50 pixels tall) and ALL (visibility is at least 20\% with at least 20 pixels tall) setups are also used to evaluate the detection performance. We use the evaluation code provided by Caltech \cite{dollar2012pedestrian} and CityPersons \cite{Shanshan2017CVPR} to obtain the corresponding MR value.  We report the runtime of proposed SSA-CNN using a single NVIDIA GTX 1080 Ti GPU.

\begin{figure}[t]
\centering
\subfigure[]{ 
\label{Caltech_Comp}
\includegraphics[width=0.85\linewidth]{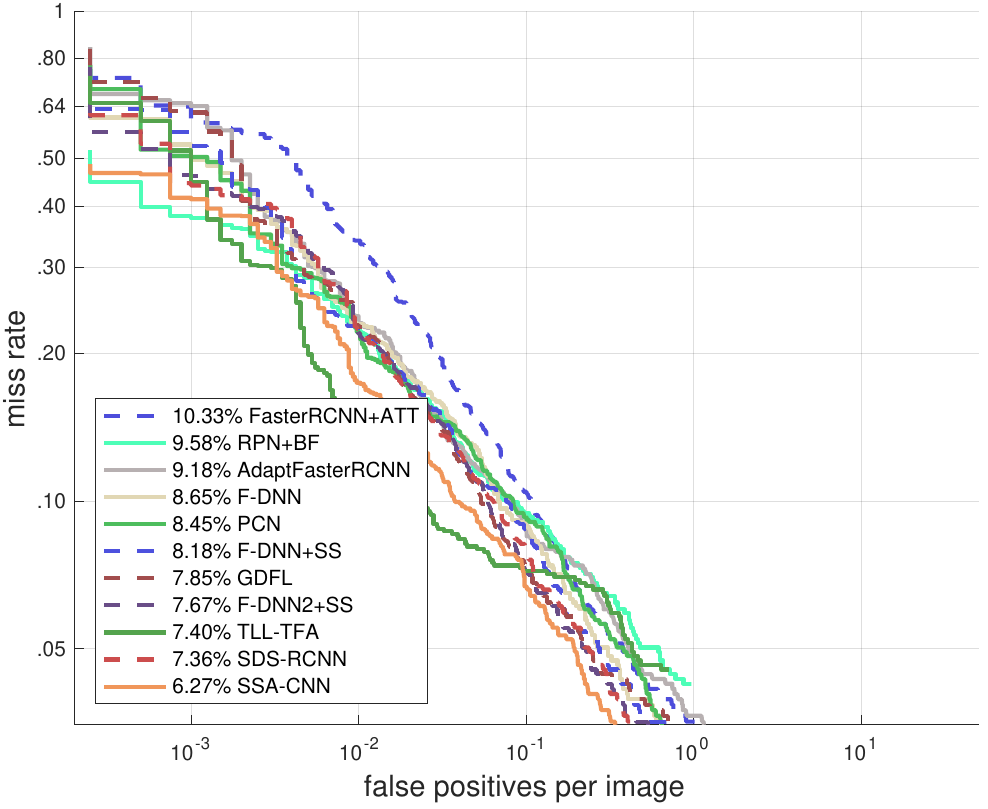}}  
\subfigure[]{ 
\label{Caltech_Comp_partial}
\includegraphics[width=0.85\linewidth]{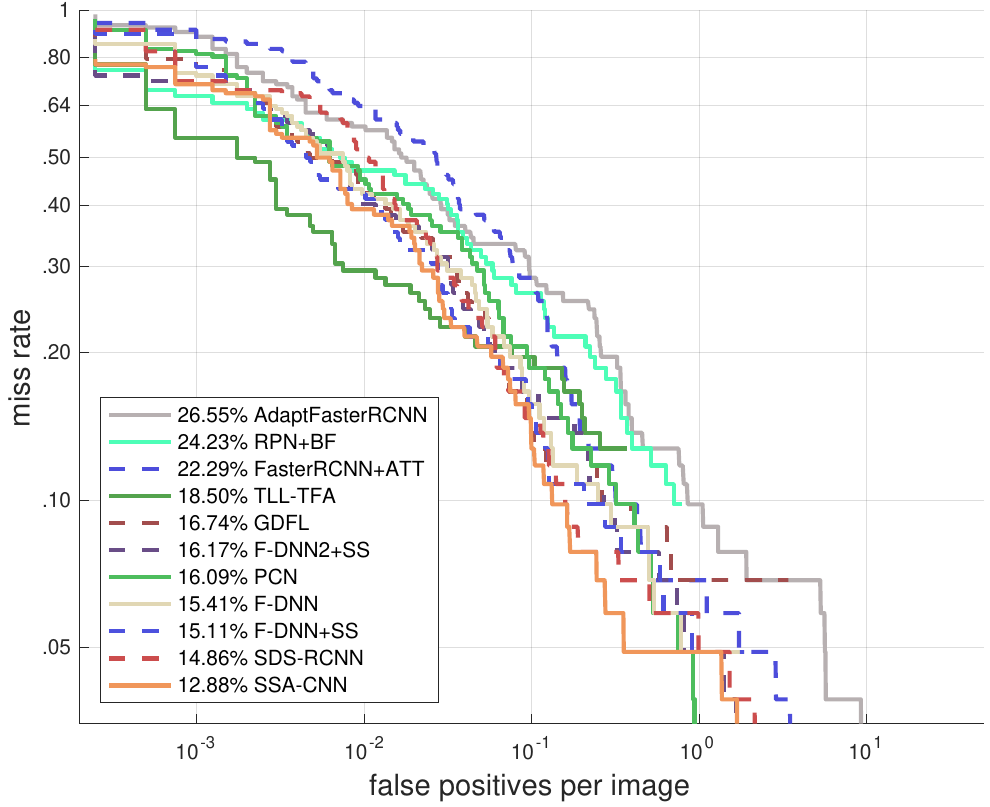}} 
\caption{Performance comparisons with state-of-the-art methods on Caltech test set under a) Reasonable  and b) Partial occlusion  setup.}
\end{figure}

\begin{table*}[t]
\caption{Performance and runtime  comparisons with state-of-the-art methods on Caltech test set under Reasonable setup. $\ast$ indicates that additional data is used when training model. }
\scriptsize
\centering

\begin{tabular}{C{1cm}  C{1.9cm} C{1.2cm} C{1.7cm} C{1cm} C{1cm} C{0.9cm} C{1.5cm} C{1.5cm} C{1.5cm}}

\cline{1-10}
Method   & CompACT-Deep & MS-CNN & SA-FastRCNN & RPN+BF & $\text{F-DNN}^{\star}$ & GDFL & $\text{F-DNN2+SS}^{\star}$ & SDS-RCNN & SSA-CNN   \\ 
\hline
MR(\%)          & 11.75        & 9.95   & 9.68        & 9.58   & 8.65  & 7.85 & 7.67      & 7.36     & 6.27      \\ 
Runtime         & 1s           & 0.4s   & 0.59s        & 0.6s    & 0.3s   & 0.05s & 2.48s      & 0.21s     & 0.11s      \\ 
GPU  &  K40      &  Titan X   &  Titan X  &   K40    &   Titan X   &  1080 Ti &  Titan X     & Titan X     & 1080 Ti      \\ \cline{1-10}
\end{tabular}
\label{caltech_dt_runtime}
\end{table*}


\begin{table}[h]
\caption{New performance comparisons with state-of-the-art methods on CityPersons dataset.}
\tiny
\centering
\begin{tabular}{C{1.1cm} C{2.0cm} C{1.0cm} C{0.7cm} C{0.6cm} C{0.6cm}}
\hline
Method                      & Configuration   & Reasonable(MR\%) &Small(MR\%) & Heavy(MR\%) & ALL(MR\%)\\ \hline
FasterRCNN                  & -                           & 15.52 &   -  & 64.83  & -   \\ \hline
\multirow{3}{*}{FasterRCNN} & channel-wise self attention & 20.93 &   -  & 58.33  & -   \\ 
                            & visible box supervision     & 16.40 &   -  & 57.31  & -   \\ 
                {+ATT}      & body part detection         & 15.96 &   -  & 56.66  & -   \\ \hline
\multirow{3}{*}{RepLoss}    & RepGT Loss                  & 13.70 &   -  & 57.50  & -   \\ 
                            & RepBox Loss                 & 13.70 &   -  & 59.10  &  -  \\ 
                            & RepGT + RepBox Loss         & 13.20 &   -  & 56.90  & -   \\ \hline
PDOE+RPN                     & -     & 11.24 & 47.35 &44.15 &43.41 \\ \hline
SSA-CNN                     & semantic self-attention     & 18.20 & 45.76& 44.31 &45.16 \\ \hline
\end{tabular}
\label{Citypersons_cmp}
\end{table}

\begin{table*}[t]
\scriptsize
\caption{Ablation experiments of RPN and R-CNN  stage evaluated on Caltech test set. \checkmark indicates corresponding component is used. The lowest MR are highlighted with bold font.}
\begin{tabular}{C{1.3cm}|C{3cm}|C{1.1cm} C{1.1cm} C{1.1cm} C{1.1cm} C{1.1cm} C{1.1cm} C{1.1cm} C{1.1cm} C{1.1cm}}
\hline
               stage        & component                    & \multicolumn{8}{c}{choice} \\ \hline
\multirow{5}{*}{RPN}   & conv5\_1\_SA &      &       &      &       &\CKM  &     &\CKM &\CKM\\                       
                       & conv5\_2\_SA &      &       &      &\CKM   &      &\CKM &     &\CKM\\
                       & conv5\_3\_SA &\CKM  &\CKM   & \CKM &       &      &\CKM &\CKM &\CKM \\  
                       & conv4\_3\_DT &      &\CKM   & \CKM &\CKM   &\CKM  &\CKM &\CKM &\CKM   \\  
                       & conv4\_3\_SA &      &       & \CKM &\CKM   &\CKM  &\CKM &\CKM &\CKM   \\ \hline
\multicolumn{2}{c|}{MR(\%)}           &9.81  &10.90  &\textbf{8.95}  &11.09  &11.42 &10.79&11.38&10.86    \\ \hline
\multirow{4}{*}{R-CNN} & conv5\_1\_SA &      &       &      &       &\CKM  &     &\CKM &\CKM\\
                       & conv5\_2\_SA &      &       &      &\CKM   &      &\CKM &     &\CKM\\
                       & conv5\_3\_SA &      & \CKM  &\CKM  &       &      &\CKM &\CKM &\CKM\\ 
                       & conv4\_3\_SA &      &       &\CKM  &\CKM   &\CKM  &\CKM &\CKM &\CKM  \\ \hline
\multicolumn{2}{c|}{MR(\%)}           &6.71  &6.79   &\textbf{6.27}  &6.78   &6.68  &6.66 &6.64 & 6.53 \\ \hline
\end{tabular}
\label{SA-CNN-ablation}
\end{table*}

\subsection{Comparisons with state-of-the-art methods on benchmarks}

\textbf{Caltech dataset:} Fig. \ref{Caltech_Comp} shows detection performance comparisons under Reasonable setup with state-of-the-art methods including FasterRCNN+ATT \cite{zhang2018occluded}, RPN+BF \cite{zhang2016faster}, AdaptFasterRCNN \cite{Shanshan2017CVPR}, PCN \cite{wang2018pcn}, F-DNN+SS \cite{du2017fused}, GDPL \cite{lin2018graininess}, F-DNN2+SS \cite{du2018fused}, TLL-TFA \cite{song2018small} and SDS-RCNN \cite{brazil2017illuminating} on Caltech test set. It can be observed that the proposed SSA-CNN achieves the best detection performance with MR of 6.27\% which is about 14.8\% improvement over current top performing SDS-RCNN \cite{brazil2017illuminating}. This performance gain is very significant considering the  detection performance on Caltech dataset is almost saturated. Compared with FasterRCNN+ATT, the proposed method achieves about 2.9\% lower on MR which implies that the proposed semantic self-attention mechanism is more robust than channel-wise attention mechanism for the task of pedestrian detection. Besides, our proposed semantic self-attention mechanism only requires box-wise annotations rather than visible-box annotations or additional human body part detector to obtain part-based semantic information as in FasterRCNN+ATT. Compared with attention mechanism proposed in GDPL \cite{lin2018graininess}, which focuses on pedestrians with small size and occlusions, the proposed method can effectively integrate semantic information for all pedestrians and achieves better overall detection performance.
Even though F-DNN2+SS \cite{du2018fused} incorporates highly accurate pixel-wise semantic information trained from Cityscapes dataset \cite{cordts2016cityscapes}, its semantic segmentation model is trained independently from the proposal generator. This means that the semantic features cannot be infused into the proposal generator. This partly explains why their detection performance is worse than our proposed method even though better semantic information is used. Fig. \ref{Caltech_Comp_partial} shows detection performance comparisons under Partial occlusion setup. The proposed method outperforms the baseline method SDS-RCNN \cite{brazil2017illuminating} by about MR of 1.98\% and obtains better performance than PCN and F-DNN+SS which work well for occluded pedestrians. 
The performance achievement demonstrates that the proposed method can effectively fuse semantic features and detection features by relying only on weak box-wise annotations. 

The detection performance and runtime comparisons with state-of-the-art methods on Caltech dataset are shown in Table \ref{caltech_dt_runtime}. From the table, it can be observed that the proposed SSA-CNN can run about 22.5X and 1.9X faster than F-DNN2+SS and SDS-RCNN respectively. Even though the GDPL \cite{lin2018graininess} obtain better computational efficiency due to small input image resolution (i.e., 640$\times$480), its MR is about 1.58\% higher than proposed method. These detection performance and runtime comparisons demonstrate that the proposed method can achieve better trade-off between performance and runtime efficiency than state-of-the-art methods and is able to work as a strong baseline for future research.

\textbf{CityPersons dataset:} The detection performance comparisons with state-of-the-art methods on CityPersons dataset are shown in Table \ref{Citypersons_cmp}. The MR under Reasonable, Small, Heavy occlusion and ALL setup of FasterRCNN+ATT \cite{zhang2018occluded}, RepLoss \cite{wang2017repulsion} and PDOE+RPN \cite{zhou2018bi} with varied components are listed. The FasterRCNN+ATT  focuses on exploring channel-wise weights as attention information to help pedestrian detection and RepLoss proposes a new loss for bounding box regression to handle the occlusions problem in the crowd case. PDOE+RPN \cite{zhou2018bi} propose an approach to simultaneous pedestrian detection and occlusion estimation by regressing two bounding boxes to localize the full body and visible part of a pedestrian respectively. 
It is worth noting that we do not use any additional information (i.e., the visibility of bounding box) except the pedestrian bounding box annotation for the task of detection in our experiments. All the results in Table \ref{Citypersons_cmp} is obtained when using original image (i.e., resolution of 2048$\times$1024) as network input. From the table, we can observe that the proposed method achieves the best detection performance under all setups.
 When evaluated on Heavy occlusion setup, the proposed method obtains much better detection performance. The MR of proposed method is much lower than FasterRCNN+ATT with channel-wise self attention and visible box supervision respectively. This further demonstrates that the proposed semantic self-attention mechanism is better than the attention mechanism used in FasterRCNN+ATT \cite{zhang2018occluded}.
 The RepLoss is designed to detect occluded pedestrian from crowd and achieves better detection performance than FasterRCNN+ATT. However, the detection performance of RepLoss are still much worse than ours, especially on Heavy occlusion setup which implies that the proposed method is a better way to handle occlusion.
 The PDOE+RPN explores the visibility annotation by regressing two bounding boxes. However, its detection performance is still worse than ours.
 These demonstrate the robustness of proposed semantic self-attention mechanism when dealing with the occlusion cases and can be explored to handle occlusions problem.

\subsection{Ablation Study}


In this subsection, we conduct ablation experiments on Caltech dataset to show the effectiveness of each component of the proposed method. We use SDS-RCNN as our baseline and progressively add each of the proposed component to evaluate their effects on the detection performance using Reasonable setup. We use conv*\_*\_SA to represent that semantic segmentation branch is added in corresponding layer and the semantic feature map is used as self-attention feature map to improve pedestrian detection in RPN and pedestrian classification in R-CNN stage. For instance, conv5\_3\_SA in RPN stage means a semantic segmentation branch is added to conv5\_3 layer and the corresponding semantic feature map is concatenated with conv5\_3 convolution feature map to serve as features for pedestrian detection. The conv4\_3\_DT means that a detection branch is connected to conv4\_3 layer. We use proposed SSA-RPN to extract pedestrian proposals when conducting ablation experiments on R-CNN stage.  Table \ref{SA-CNN-ablation} summarizes the ablation experiments.

\textbf{Semantic Self-Attention in RPN}: We first analyse the influence of each component in RPN stage. From the upper part of Table \ref{SA-CNN-ablation}, we can observe that MR decrease from 9.81\% to 8.95\% when adding detection and semantic attention branches to conv4\_3 layer which implies that the multi-scale detection and semantic segmentation have positive effects on the detection performance. It is also evident that the detection performance deteriorates when semantic self-attention is removed in conv4\_3 layer which means that
the multi-scale semantic segmentation plays an important role to boost the detection performance. In addition to using semantic feature map from conv5\_3 layer to serve as self-attention feature map, we also examine the detection performance when using other conv5 layers. From Table \ref{SA-CNN-ablation}, we can see that the detection performance drops significantly when semantic feature maps from conv5\_1 and conv5\_2 layers are used, especially when semantic feature map from conv5\_1 layer is employed as part of self-attention feature maps. This means that the quality of semantic attention information has huge impact on the detection performance and semantic feature maps from deeper layer can provide better semantic self-attention feature maps.

\textbf{Semantic Self-Attention in R-CNN}: From the lower part of Table \ref{SA-CNN-ablation}, we can observe that the MR decreases from 7.36\% in SDS-RCNN to 6.71\% when using proposed SSA-RPN to extract pedestrian proposals and R-CNN from SDS-RCNN to refine the proposals. This demonstrates that the proposed SSA-RPN can extract better pedestrian proposals than SDS-RCNN and the quality of proposals have huge influence for final detection performance. It can be observed that the MR further decreases from 6.78\% to 6.27\% when using proposed multi-scale semantic self-attention mechanism, where the semantic attention information from conv4\_3 and conv5\_3 layer are exploited.  This demonstrates the effectiveness of the proposed multi-scale semantic self-attention infusion mechanism. Similar with RPN stage, we fix the semantic feature map from conv4\_3 and check the influence of semantic self-attention information from other conv5 layers. From Table \ref{SA-CNN-ablation}, we observe that the detection performance fluctuates when semantic information from other conv5 layers are used. These results highlight that the semantic feature maps from deeper network layers provide more robust semantic information and play a key role in achieving higher detection performance.


\section{Conclusion}
\label{conclusion}

In this work, we propose an approach to explore semantic segmentation as self-attention to improve pedestrian detection. The semantic feature maps obtained using semantic segmentation served as self-attention information and are concatenated with corresponding convolution feature maps to provide more discriminative features for pedestrian detection in RPN and pedestrian classification in R-CNN. We further incorporated pedestrian detection and semantic segmentation in different scales to boost the detection performance in RPN stage and observe that the feature maps become more discriminative with the proposed method. We integrate the semantic segmentation obtained from multi-scale network layers to serve as self-attention information in R-CNN stage and demonstrate that considerable performance improvement is achieved with the proposed semantic self-attention mechanism. We report experiment results on well-known Caltech dataset and provide a new state-of-the-art detection performance with MR of 6.27\% which is about 14\% improvement over the currently reported best method.

{\small

}

\end{document}